\documentclass[10pt,twocolumn,letterpaper]{article}

\usepackage{iccv}
\usepackage{times}
\usepackage{epsfig}
\usepackage{graphicx}
\usepackage{amsmath}
\usepackage{amssymb}

\newtheorem{theorem}{Theorem}
\newtheorem{lemma}[theorem]{Lemma}
\usepackage[pagebackref=true,breaklinks=true,letterpaper=true,colorlinks,bookmarks=false]{hyperref}

\iccvfinalcopy % *** Uncomment this line for the final submission

\def\iccvPaperID{****} % *** Enter the ICCV Paper ID here

\ificcvfinal\fi

\begin{document}

%%%%%%%%% TITLE
\title{Instance-wise Linearization of Neural Network for
Model Interpretation}

\author{Zhimin Li\\
University of Utah\\
{\tt\small zhimin@sci.utah.edu}
\and
Shusen Liu\\
Lawrence Livermore National Laboratory\\
{\tt\small liu42@llnl.gov}
\and
Bhavya Kailkhura\\
Lawrence Livermore National Laboratory\\
{\tt\small kailkhura1@llnl.gov}
\and
Peer-Timo Bremer\\
Lawrence Livermore National Laboratory\\
{\tt\small bremer5@llnl.gov}
\and
Valerio Pascucci\\
University of Utah\\
{\tt\small pascucci@sci.utah.edu}
}

\maketitle
\ificcvfinal\thispagestyle{empty}\fi

%%%%%%%%% ABSTRACT
\begin{abstract}
Neural network have achieved remarkable successes in many scientific fields.
However, the interpretability of the neural network model is still a major bottlenecks to deploy such technique into our daily life.
The challenge can dive into the non-linear behavior of the neural network, which rises a critical question that how a model use input feature to make a decision. 
The classical approach to address this challenge is feature attribution, which assigns an important score to each input feature and reveal its importance of current prediction.
However, current feature attribution approaches often indicate the importance of each input feature without detail of how they are actually processed by a model internally.
These attribution approaches often raise a concern that whether they highlight correct features for a model prediction.

For a neural network model, the non-linear behavior is often caused by non-linear activation units of a model.
However, the computation behavior of a prediction from a neural network model is locally linear, because one prediction has only one activation pattern.
Base on the observation, we propose an instance-wise linearization approach to reformulates the forward computation process of a neural network prediction.
This approach reformulates different layers of convolution neural networks into linear matrix multiplication.
Aggregating all layers' computation, a prediction complex convolution neural network operations can be described as a linear matrix multiplication  $F(x) = W \cdot x + b$.
This equation can not only provides a feature attribution map that highlights the important of the input features but also tells how each input feature contributes to a prediction exactly.
Furthermore, we discuss the application of this technique in both supervise classification and unsupervised neural network learning  parametric t-SNE dimension reduction.
\end{abstract}

%%%%%%%%% BODY TEXT
\section{Introduction}
Neural network techniques have achieved remarkable success across many scientific fields~\cite{jumper2021highly, senior2020improved, silver2016mastering, esteva2017dermatologist}.
Furthermore, many applications (e.g., autopilot, debt loan, cancer detection, criminal justice,...) which utilize this technology, starts to involve into our daily life and even more in the expected future.
However, the uninterpretable behavior of the neural network prediction causes many concerns.
For example, recently, a few countries' governments have published AI act~\cite{the_white_house_2022,veale2021demystifying} to regulate AI systems and require automate systems must provide an explanation for why it makes such a decision. 
Meanwhile, improving the interpretability of the neural network can also benefit scientific fields for knowledge discover, such as drug discover~\cite{vamathevan2019applications,artrith2021best} 

A key challenge of the neural network interpretability is the non-linear behavior of the neural network model.
These non-linear behaviors are caused by different activation patterns which are triggered by different inputs, and these diverse activation patterns lead to an uninterpretable behavior.
A classical approach to explain neural network prediction is the feature attribution method~\cite{lundberg2017unified, ribeiro2016should}, which assigns an important score to each input feature and highlight the most important features.
However, different feature attribution approaches may end up with different feature maps, and how input features are processed by the neural network model internally is mysterious~\cite{zhou2022feature}.

\begin{figure}[t]
\centering
    \includegraphics[width=\linewidth]{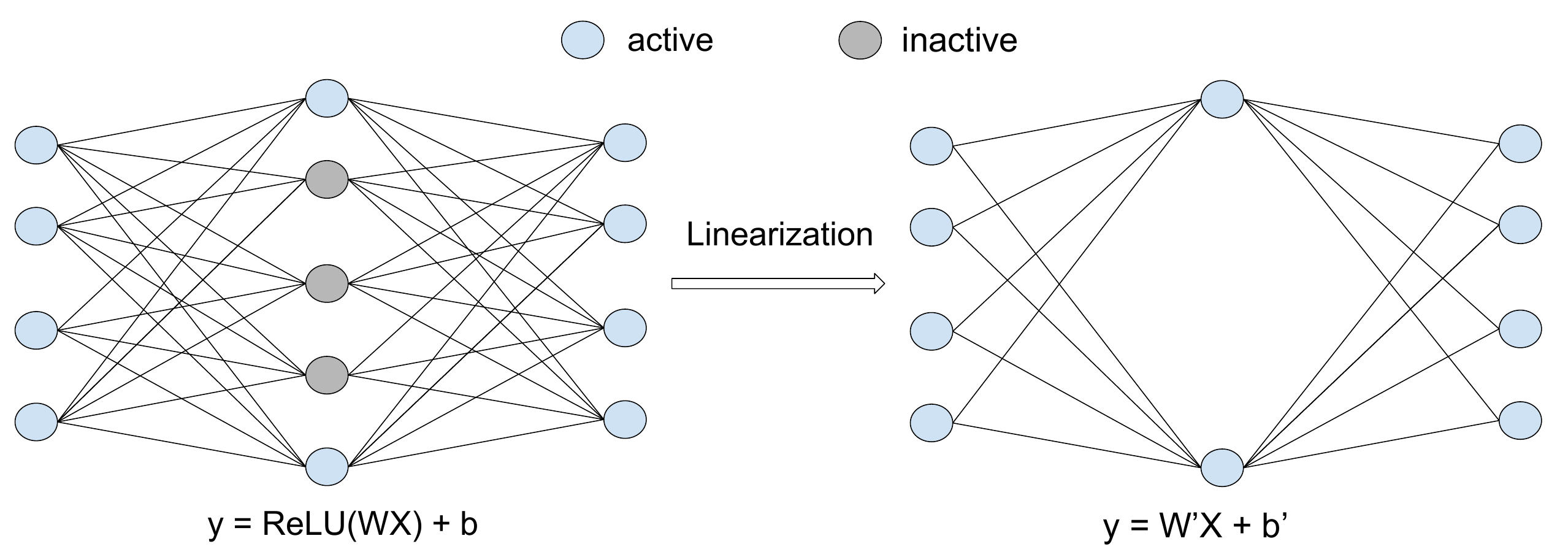}
    \vspace{-3mm}
    \caption{A prediction of neural network model has only one activation pattern and the complex prediction operation can be simplified as a linear matrix multiplication.}
\label{fig:linearization}
\vspace{-2mm}
\end{figure}

For a single neural network prediction, the neural network model has only single activation pattern. 
The input features which are processed by the neural network layer by layer during the prediction is linear.
It raises a question that whether the computation process of a neural network can generate an answer of how each input feature contributes to the prediction. 
In this study, we propose an approach to reformulate the decision process of a neural network prediction and reformulate the forward computation process of a neural network prediction.
Our approach is based on an observation that
a neural network prediction has only one activation pattern.
As the Figure~\ref{fig:linearization} shows, the non-linear ReLU activation components of a neural network prediction can be consider as linear components.
Therefore, the non-linear matrix multiplication of a neural network performs on a single prediction is linear process.
The whole computation process can be reformulated and aggregated into a linear equation $F(x) = W\cdot x + b$.

We find that this linear equation can be used to explain how the input feature is used by the neural network model to make prediction.
In this paper, we discuss how to reformulate the forward propagation of a neural network layer by layer and discuss the potential challenge of this computation.
We propose an efficient approach to calculate $W$, $b$, and discuss the role of $W$ and $b$ during model prediction.
Our approach is much more flexible and can be applied to supervise learning task or unsupervised learning task.
At the end, we demonstrate a use case of how neural network models capture input feature in different layer of network models and how researchers can use our technique to understand the behavior of neural network models.

%\begin{itemize}
%    \item A novel framework to reformulate the forward computation of a neural network prediction and represent the prediction as a local linear equation(sec. 3.1-3.6).

%    \item A feature attribution approach that reveals the contribution of each input feature to a neural network model prediction and the activation of a neuron exactly (sec. 3.7-3.8).

%    \item A comparison study to compare our approach of state of art feature attribution approach and a evaluation demonstrate the usability of our approach and highlight the potential challenges (sec. 4, 5).    
%\end{itemize}

\section{Relative Work}
Because of the black-box nature of the neural network model, understanding how it makes prediction is an important but challenge topic.
In this section, we discuss previous techniques that have been proposed in the literature to address this challenge.

\subsection{Feature Attribution}
Feature attribution is a classical approach, which generates a heat map to highlight the importance of input feature for a model prediction. 
Different methodology has been designed to calculate the heat map.
Gradient base approaches~\cite{simonyan2013deep,sundararajan2017axiomatic,selvaraju2017grad,smilkov2017smoothgrad} calculate the model gradient of an input with respect to the target label and use or accumulate the gradient values to highlight the importance of each input feature.
Perturbation base approaches~\cite{petsiuk2018rise,fong2019understanding,fong2017interpretable,zeiler2014visualizing,ribeiro2016should} ablate or modify a part of the input and observe the output variation to understand the contribution of each input feature to the prediction.
Other approach like SHAP~\cite{lundberg2017unified}, Deep-Lift~\cite{shrikumar2017learning}, and LRP~\cite{bach2015pixel} provide a feature attribution map from different angle.
However, because of the lacking of ground truth, whether a feature attribution approach highlights the real important regions of an input is still under exploration. 

Understanding features that captured by a neuron of neural network can also improve the interpretability of neural network. 
Feature visualization~\cite{olah2017feature} optimizes an input image by maximizing a neuron's activation value. 
The output of the optimization provides information about what features are captured by the neuron.
Previous researchers~\cite{bau2017network,zhou2014object} have also measured the alignment between individual neuron unit and semantic concepts.
Fong et al.~\cite{fong2019understanding} applies input perturbation to measure the reaction of a neuron to these perturbation and capture the input regions that contribute most to these activation.

\subsection{Local Linearity of Neural Network}
Researchers have investigated the local linearity of ReLU network, which mainly focus on the complexity of the model such as approximating the number of linear regions~\cite{robinson2019dissecting,serra2018bounding,hanin2019complexity,montufar2014number,raghu2017expressive,serra2020empirical}.
Previous researches have also aligned local linearity with input perturbation to understand a model's prediction robustness and generalization.
Novak, et al~\cite{novak2018sensitivity} performs a large scale experiments on different neural network models to show that input-output Jacobian norm of samples is correlated with model generalization.
Lee, et al.~\cite{lee2019towards} design algorithm to expand the local linear region of neural network model.

%Comparing with previous work, our study shows that for each model prediction, we can construct a local linear equation that map the input sample to output prediction, and we show how to calculate the linear equation to explain the network prediction.
%\cite{ortiz2021can}
\section{Instance-wise Linearization}
%Linear transformation is critical to improve the interpretability of a neural network model. 
%The challenging of the linear transformation is the diverse neuron activation patterns of model with different inputs which is often interpreted as non-linear behavior of a neural network model.
%However, instead of  constructing a linear approximation the complex activation pattern to explain the neural network prediction globally, a single input has only one activation pattern and the prediction behavior of a model on it is locally linear.

In this section, we discuss how to transform different layers of neural network into linear operation under this condition.
After transforming all layers, we demonstrate how to aggregate all linear operations into a linear matrix multiplication (1).
In this study, we mainly focus our discussion on convolution neural network, which is the de-faco setting for many computer vision tasks.
\begin{align}
F(x) = W \cdot x + b
\end{align}

A series operations of a neural network can be represented as a nest function $F(x) = f_{n}\cdot f_{n-1}....f_{2}\cdot f_{1}$ and n is the number of layers in neural network.
The detail operation of each layer can be generalized as
$f_{i} = \sigma(W_i\cdot x_i + b_i)$.
$W_i$, $x_i$ and $b_i$ are the weight, feature representation and bias of i-th layer. $\sigma$ is the activation function of neural network model.
%During our discussion, we assume that the activation function $\sigma$ used in the neural network is ReLU (or piece-wise linear function) and the neural network processes one input sample at a time. 

\subsection{Fully Connected Layer}
To explain above statement, we use a fully connected layer as an example.
Our discussion mainly focus on the activation functions  such as ReLU, GELU, SELU and ELU.
\begin{align}    
y = \sigma(W_{i} \cdot x_{i}+ b_{i}) \label{eq:0}
\end{align}

$W_{i} \cdot x_{i}$ is the dot product operation. $x_i$ is the feature vector, and $\sigma$ is the activation function that changes the computation output.
For each activation function, the computation can be transformed to linear matrix multiplication case by case, and the equation (2) is rephrased as 
$y = \lambda (W_{i} \cdot x_{i}+ b_{i})$.
Here, the variable $\lambda$ is a re-scale factor which represents the impact of activation function on the computation result.

%$M$ is the indicator matrix which consists zero or one that represents the weight corresponds to a certain neuron is activated or not. 
%$M$ can be trivially constructed from vector $m$.
%$W \odot M$ is the element-wise product, $W^{'} = W \odot M$, and
%$b^{'} = b \odot m$.
%The convolution layer can be considered as a sparse fully connected layer and transformed into the linear matrix multiplication. Similarly, the maximum pooling layer and skip connection layer can also be transformed into $y = W^{'}\cdot x + b^{'}$ (see details in support material). 

\subsection{Convolution Layer}
The convolution layer can be considered as a sparse fully connected layer and transformed into the linear matrix multiplication.
Assume the kernel size is (c, k, k), the stride size is s, and the input tensor with a size $channel \times width \times height$ can be flatten as a channel * width * height dimension vector $x$.
Each output element of a convolution operation can be considered as a dot product of vector multiplication $w_i \cdot x + b_i$. 
Here, $w_i$ is the same size as the vector $X$ and most of elements are zero except elements which are operated by the kernel operation. 
The overall convolution operation can be converted into $W \cdot x +b$. In the equation, $W$ is the toeplitz matrix which is consisted of $w_i$, and $b$ is consisted of $b_i$.
The convolution operation also comes with the activation function, so overall equation is $ReLU(W \cdot x +b)$.
Similar to equation (1) and (2), can be rephrase above equation  as $W^{'} \cdot x + b^{'}$. %$M$ and $m$ are the activation matrix and vector.

\begin{figure}[htp]
\centering
    \includegraphics[width=\linewidth]{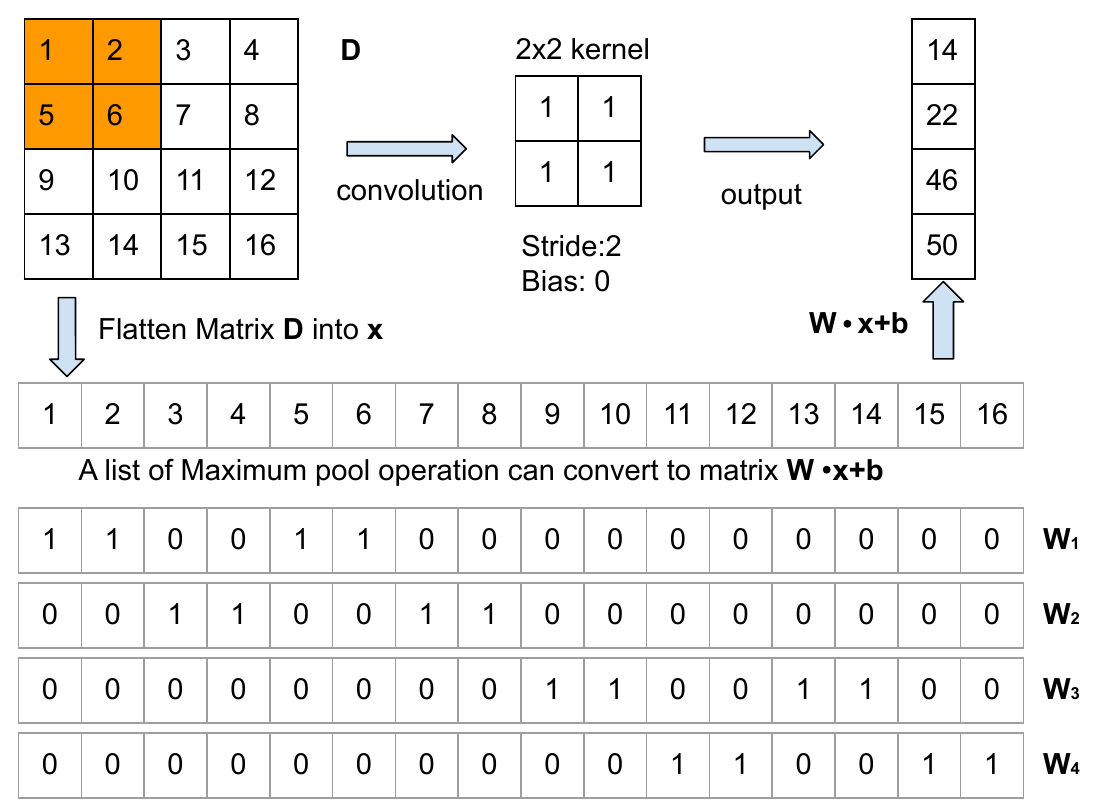}
    \vspace{-3mm}
    \caption{A convolution operation can be rephrased as a matrix vector multiplication. Performing a 2x2 convolution operation with stride size 2 on a 4x4 input matrix can be rephrased as a dot product between matrix and input vector $W\cdot x + b$.}
\label{fig:convolution2linear}
\vspace{-2mm}
\end{figure}

In Figure~\ref{fig:convolution2linear}, we demonstrate how to transform a 2x2 convolution kernel with stride 2 on a 4x4 matrix input into a dot product between a matrix and a vector. The bias term is set to zero.
The original 4x4 matrix can be flatten into a 16 dimension vector \textit{a}.
The first convolution kernel operation on the matrix x can be converted into a vector $w_1$ and the overall convolution operation is converted into a matrix $W$.
%Similar to the convolution layer, maximum pooling, skip connection, and batch normalization can also convert into $y = W^{'}\cdot x + b^{'}$. 
%The detail is described in the support material.

\subsection{Pooling Layer}
The pooling operation in the neural network can  be describe as matrix multiplication $W \cdot x$.
Assume the pooling kernel size is (c, k, k), the stride size is s, and the input tensor with a size $channel \times height \times width$ can be flatten as a channel*height*width dimension vector $x$.
We use the maximum pooling as an example to discuss transformation process and the other pooling (e.g., average pooling) operations should be similar.
An element of output tensor is a dot product of indicator vector $W_{(i,j,k)}$ (i,j,k is the index of the output tensor) indicates the elements of input $x$ that are selected by the maximum pool operation. 
Therefore, the pooling operation can be rephrased as $W \cdot x$.

\begin{figure}[htp]
\centering
    \includegraphics[width=\linewidth]{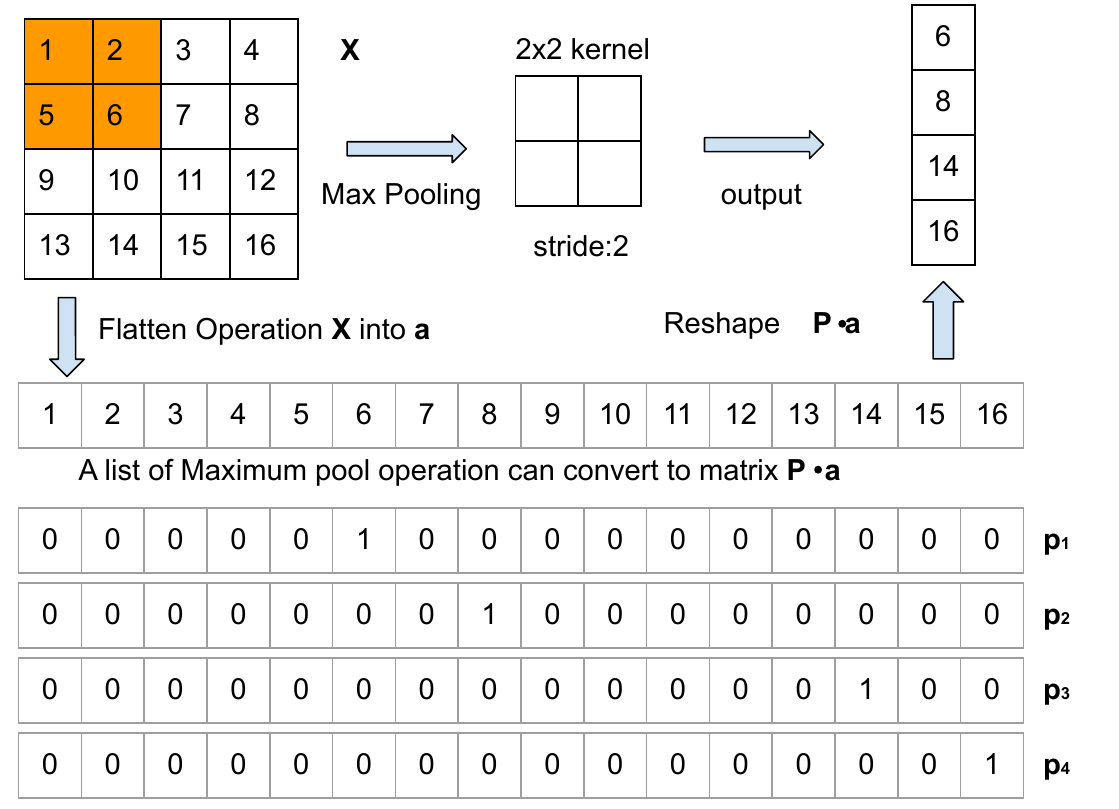}
    \vspace{-3mm}
    \caption{A maximum pooling operation can be rephrased as a matrix vector multiplication. Performing a 2x2 maximum pooling operation with stride size 2 on a 4x4 input matrix can be rephrased as $W\cdot a$}
\label{fig:maxpooling2linear}
\vspace{-2mm}
\end{figure}

In Figure~\ref{fig:maxpooling2linear} demonstrate a case which transforms a 2x2 maximum pooling operation with stride 2 on a 4x4 matrix into a dot product between a matrix and a vector.
The original 4x4 matrix can be flatten into a 16 dimension vector \textbf{x}.
The first maximum pooling operation on the matrix X can be convert into a vector $w_1$ and the overall maximum pooling operation is converted into a matrix $W$,

\subsection{Skip Connection}
\begin{figure}
\centering
    \includegraphics[width=\linewidth]{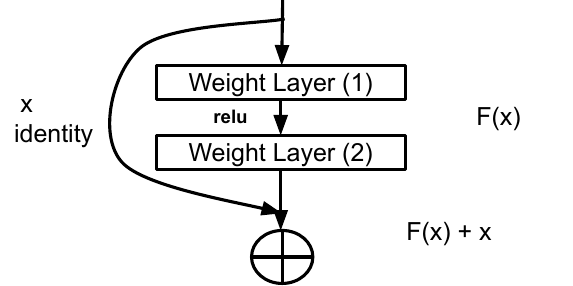}
    \vspace{-3mm}
    \caption{A skip connection component in ResNet architecture.}
\label{fig:convolution_resnet}
\vspace{-2mm}
\end{figure}

Skip connection is a critical component in the ResNet~\cite{he2016deep} architecture.
In Fig.~\ref{fig:convolution_resnet} is an example of the skip connection and the equation can be representation as $F(x) + x$. In the equation, $F(x)$ is often consisted of convolution layer, batch normalization, ReLU activation and Full connected layers.
As we discuss in previous section, these layers can be simplified as a linear matrix multiplication
$F(x) = W_{skip}\cdot x + b_{skip}$. The overall equation can be rephrase as 
$$(W_{skip}+ I)\cdot x + b_{skip}$$
Merging the identify matrix $I$ into $W_{skip}$ and the equation can be simplified into $W^{'}_{skip} \cdot x + b_{skip}$.

\subsection{Batch Normalization Layer}
Batch normalization~\cite{ioffe2015batch} is a critical component in neural network model which makes the training easy and efficient.
The inference process of the batch normalization layer is an element-wise linear transformation in equation~\ref{eq:batch}.
The equation that used to perform batch normalization on a single value is described as following:

\begin{align}
y = \frac{x_i - \mu}{\sqrt{\sigma^2 + \epsilon}} * \gamma + \beta = \frac{\gamma}{\sqrt{\sigma^2 + \epsilon}} * x_{i} + \frac{-\mu \gamma}{\sqrt{\sigma^2 + \epsilon}} +\beta \label{eq:batch}
\end{align}

In the equation, $\mu$, $\sigma^2$, $\gamma$, and $\beta$ are calculated during the training process. These variables are available during the network inference.
This equation can be rephrased as $y = w_i^{norm} * x_i +  b_{i}^{norm}$
and $w_i^{norm} = \frac{\gamma}{\sqrt{\sigma^2 + \epsilon}},\; b_{i}^{norm} = \frac{-\mu \gamma}{\sqrt{\sigma^2 + \epsilon}} +\beta$.
The batch layer operation can be represented as:
$$W^{norm} \odot x +b$$

\subsection{Layer Aggregation}
We have mentioned that a neural network can be defined as a nest function
$F(x) = f_{n}\cdot f_{n-1}....f_{2}\cdot f_{1}$ and each layer's function can be generalized as following:
$$f_{i} = \sigma(W_i\cdot x_i + b_i)$$

With our previous description, for a single prediction, each layer can be replaced as a linear matrix multiplication  
$$f_{i} = W_i^{'} \cdot x_i + b_i^{'}$$
and the overall equation can be rephrased as
$$
F(x) = W_{n}^{'} \cdot W_{n-1}^{'} ... W_{2}^{'} \cdot W_{1}^{'} \cdot x + \sum_{i=1}^{n-1
}(W_{n}^{'}...W_{i+1}^{'} \cdot b_{i})+b_{n}
$$

After linearization of a neural network prediction, the prediction is represented as $F(x) = W \cdot x + b$. In the equation, $W = W_{n}^{'} \cdot W_{n-1}^{'}\cdot... W_{2}^{'} \cdot W_{1}^{'}$ and $b = \sum_{i=1}^{n-1}(W_{n}^{'}...W_{i+1}^{'} \cdot b_{i})+b_{n}$.
For piece-wise linear activation function, the inference behavior of neural network model is locally linear.
In a local region, the behavior of neural network on input x is equal to $F(x) = W \cdot x + b$. 

%$F(x)$ is the output of the original model which can be computed directly. To calculate the result of $W^{*} \cdot x$, we can set all $b_i$ in $b^{*}$ to zero such that $b^{*} = 0$ and $F_{b^{*}=0}(x) = W^{*}\cdot x$.
%The final term $b^{*} = F(x) - W^{*} \cdot x$. During the computation, it is important to notice that $W_i$ in each layer should maintain the same (e.g., same activation, and maximum pooling location).
%However, calculating the $W^{*}$ by reformulating the forward propagation is computation expensive.
%$W^{*}$ and $b^{*}$ are determined by the weights of the neural network model and the activation pattern caused by the input $x$.

\begin{lemma}
For a network with piece-wise linear activation function, for $x \in R^{m}$, there is a maximum $|\delta| \in R^{m}$ such that $\forall \theta, |\theta| < |\delta|$ and $x+\theta$ has the same activation pattern as $x$.
The matrix $W$ can be used to explain the prediction of $x+\theta$ and $x$.
\end{lemma}

Under this condition, the feature attribution $W$ is equal to input-output Jacobian matrix $J_{F}(x)$, and $W$ not only tells the sensitivity of input with respect to the model prediction but also represents how neural network use input to make prediction exactly.
ReLU networks split the input space into linear regions~\cite{montufar2014number}.  
In each linear region, these samples share the same explainable matrix $W$.
$$
J_{F}(x) = [\frac{\partial F(x)_1}{\partial x}, \frac{\partial F(x)_2}{\partial x},...,\frac{\partial F(x)_n}{\partial x}]^{T}= W
$$

For piece-wise linear activation function, a prediction of neural network has an equivalent input and output linear mapping $y=W\cdot x + b$. 
However, activation function such as GELU, SELU and ELU can not use Jacobian matrix to calculate $W$. 
Therefore, the equation for $y = W\cdot x + b$ need to be calculated layer by layer.

\subsection{Ensemble Model}

A prediction of ensemble neural network model can be represented as $$F(x) = \sum_{i=1}^{n}a_{i}F_{1}(X)$$.
$F_{i}$ is the individual neural network and the $a_{i}$ is the share assigned to the prediction of the $i_{th}$ neural network.
As we discuss in section 3.8, the equation can be rephrased as $$F(x) = \sum_{i=1}^{n}a_{i}(W_{i}^{*} \cdot x+ b_{i}^{*}) = (\sum_{i=1}^{n}a_{i}W_{i}^{*})\cdot x + \sum_{i=1}^{n}a_{i}b_{i}^{*}$$

\section{Experiment} 
The neural network prediction can be represented as a linear matrix multiplication $F(x) = W \cdot x + b$.
How do $W \cdot x$ and $b$ impact the decision of the neural network prediction is important to interpreted the decision of the network model.
To evaluate the impact of these two terms in a network prediction, we perform experiments on multiple neural network architectures that trained with MNIST, cifar10, cifar100, and Imagenet datasets.

\begin{table}[t]
\centering
\begin{tabular}{ccccc}
 \hline
Network   & $W\cdot x$ & $b_{LFR}$ & $b$ & Accuracy  \\  \hline  
    Lenet300 & 0.9794 & 0.8161 & 0.1828 &  0.982  \\
    Lenet5  & 0.9927 & 0.8408 & 0.1596 & 0.9924\\\hline
\end{tabular}
\caption{Compare the prediction accuracy and label flip rate of $W \cdot x$ and $b$ with original accuracy of MNIST test dataset over LeNet5 and LeNet300 model.}
\label{tab:mnist_result}
\end{table}

\begin{table}[t]
\centering
\begin{tabular}{ccccc}
 \hline
Network   & $W\cdot x$ & $b_{LFR}$ & $b$ & Accuracy \\  \hline    
    LeNet5  &  0.3672 & 0.6222 & 0.3307 & 0.754 \\
    AlexNet  &  0.305 & 0.3416 & 0.6223 & 0.8468 \\
    VGG16 &   0.3782 & 0.1116 & 0.8567 & 0.9123 \\
    VGG19  & 0.4257 & 0.1129 & 0.8544 & 0.9113 \\
    ResNet18  & 0.1817 & 0.0976 & 0.8756 & 0.9219 \\
    ResNet50  & 0.2496 & 0.1093 & 0.8687 & 0.9217 \\
    ResNet152  & 0.2398 & 0.0965 & 0.8758 & 0.927 \\
    DenseNet121  & 0.1957 & 0.0893 &  0.8876  & 0.9328 \\\hline
\end{tabular}
\caption{Compare the prediction accuracy and label flip rate of $W \cdot x$ and $b$ with original accuracy of cifar10 test dataset over different CNN architectures.}
\label{tab:cifar10_result}
\end{table}

\begin{table}[t]
\centering
\begin{tabular}{cccccc}
 \hline
Network   & $W\cdot x$ & $b_{LFR}$ & $b$ & Accuracy \\  \hline    
    AlexNet  &  0.0487 & 0.6504 & 0.3061 & 0.6108 \\
    VGG16  &  0.0471 & 0.3559 & 0.5918 & 0.7254 \\
    VGG19  & 0.0397 & 0.3297 & 0.6073 & 0.7176 \\
    ResNet18  &  0.1344 & 0.355 & 0.5931 & 0.7636 \\
    ResNet50  & 0.1581 & 0.3164 & 0.633 & 0.7911 \\
    ResNet101 & 0.1388 & 0.2854 & 0.6624 & 0.7961 \\
    ResNet152  & 0.1545 & 0.2898 & 0.6528  & 0.7954 \\
    DenseNet121 &  0.1553 & 0.3381 & 0.6193  & 0.7906 \\\hline
\end{tabular}
\caption{Compare the prediction accuracy and label flip rate of $W \cdot x$ and $b$ with original accuracy of cifar100 test dataset over different CNN architectures.}
\label{tab:cifar100_result}
\end{table}

\begin{table}[t]
\centering
\begin{tabular}{ccccc}
 \hline
Network  & $W\cdot x$ & $b_{LFR}$ & $b$ & Accuracy \\  \hline 
ResNet18  &  0.032 & 0.4227 & 0.5029 & 0.7028 \\
ResNet50  &  0.056 & 0.3459 & 0.5889 & 0.7673 \\
ResNet152 &  0.0632 &  0.3031 & 0.6354 & 0.7866 \\\hline
\end{tabular}
\caption{Compare the prediction accuracy and label flip rate of $W \cdot x$ and $b$ with original accuracy of 10000 samples of ImageNet validation dataset over different CNN architectures.}
\label{tab:cifar100_result}
\end{table}

%\begin{table*}[t]
%\centering
%\begin{tabular}{cccccccccc}
% \hline
%Network  & $W^{*}x_{LFR}$ Acc@1 & $W^{*}x$ Acc@1 & $W^{*}x$. Acc.@5 & $b^{*}_{LFR}$ Acc@1 & $b^{*}$ Acc@1 &  Acc@1 & Acc@5\\  %\hline  
    
    %VGG16 & 0.87492 & 0.1151 & 0.2355 & 0.1002 &  0.70372 & 0.7159 & 0.9038\\
    %VGG19 & 0.74906 & 0.2289 & 0.4053 & 0.07686 & 0.7146 & 0.7238 & 0.9088\\
    %VGG11\_bn & 0.99836 & 0.150 & 0.656 & 0.05618 & 0.69844  & 0.7037 & 0.8981 \\
    %VGG13\_bn & 0.99932 & 0.096 & 0.522 & 0.08026 & 0.70558  & 0.71586 & 0.90374\\
    %VGG16\_bn & 0.99896 & 0.120 &  0.534 & 0.00716 & 0.73366 & 0.7336 & 0.9152\\
    %VGG19\_bn & 0.99896 & 0.100 & 0.530 & 0.0279 & 0.74146  & 0.7422 & 0.9184\\
    %resnet18 & 0.99854 & 0.124 & 0.484 & 0.09138 & 0.6852 & 0.6976 & 0.8908\\
    %resnet50 & 0.9992 & 0.108 & 0.462 & 0.215 & 0.688 & 0.7613 & 0.9286 \\
    %resnet152 & 0.9988 & 0.108 & 0.540 & 0.24644 & 0.6771 & 0.7831 & 0.9405 \\
    %densenet121 &  0.99832 & 0.184 &  0.824 & 0.0456 & 0.74084 & 0.7443 & 0.9197  \\
%    \hline
%\end{tabular}
%\caption{Compare the prediction accuracy and label flip rate of $W^{*} \cdot x$ and $b^{*}$ with original accuracy of ImageNet %validation dataset over different CNN architectures.}
%\label{tab:imagenet_result}
%\vspace{-2mm}
%\end{table*}

\subsection{Decompose the Model Prediction}
Previous section has mentioned that a prediction is described as $F(X) = W \cdot x + b $.
To evaluate the impact of these two terms, we compare the original prediction accuracy of neural network $F(x)$ with the prediction of $W\cdot x$ and $b$.
%If the model without the bias term still makes the same prediction, then the matrix $W^*$ should have significant explainable power to tells how the neural network make decision with current input.
%Since bias term can not change the prediction outcome, the impact from %the $W^{*}$  is the main factor of the decision.
During the experiment, we use label flip rate (LFR) to track the prediction difference between the select term and the original prediction.
The label flip rate is defined as the number of prediction change during the decision process. 
A smaller LFR value indicates a similar prediction result with the original model.
In the experiment, we use stochastic gradient decent(SGD) to train the neural network model.
For the MNIST dataset, each model is trained with 50 epochs.
The neural network models used for cifar10 and cifar100 are trained with 200 epochs. 
We use the pre-train models from Pytorch to evaluate the ImageNet dataset.

LeNet300 and LeNet5 are the classical models that used to train MNIST dataset. Both models are trained without batch normalization layer.
In Table~\ref{tab:mnist_result}, we compare two models' prediction results with $b$ and $W \cdot x$.
The $W \cdot x$ is model accuracy without bias term and
$W \cdot x_{LFR}$ is the label flip rate compared with \textit{accuracy} which is the original accuracy of the model. 
From the evaluation result, we can tell that after removing the bias term, the label flip rates  $W \cdot x_{LFR}$ are 0.0082 and 0.0012. The $W \cdot x$ accuracy and  original accuracy is similar.
In the other hand, the impact of $b$ is an insignificant part of the prediction and the $b_{LFR}$ is large.

However, the observation from MNIST dataset does not generalize to large models and complex datasets.
In cifar10, and cifar100 dataset, we use popular convolution architectures (VGG~\cite{simonyan2014very}, ResNet~\cite{he2016deep}, and DeseNet~\cite{huang2017densely}) to compare the performance of $W \cdot x$ and $b$.
From  the results of Table~\ref{tab:cifar10_result} and Table~\ref{tab:cifar100_result}, we can tell that bias term $b$ dominates the prediction behavior of a neural network model.
Except the network such as LeNet5 and AlexNet, which have relative large impact in both $b$ and $W \cdot x$, the rest of model shows that $b$ has the dominant impact during the model prediction.
Using the information from $b$ can determine the majority prediction of the model.
In the other hand, the impact of $W \cdot x$ along is not enough to determine the model prediction.

%For model interpretation, it is common to visualize feature attribution map multiply with input vector to explain the prediction of neural network model.
%In Figure~\ref{fig:imagenet_evaluation_result}, we demonstrate a visualization result of $W^{*} \cdot x$ to explain the prediction of VGG16 from Imagenet.
%This is a common approach to explain the prediction of neural network model and $W^{*}$ tells how the neural network process the input feature.
%However, the evaluation result of VGG16 in the the ImageNet dataset (Table ~\ref{tab:imagenet_result}) tells that the prediction result of $W^{*} \cdot x$ is insignificant to explain most of the model prediction.

%\begin{figure}[t]
%\centering
%    \includegraphics[width=\linewidth]%{images/5_imagenet_evaluation.pdf}
%    \vspace{-3mm}
%    \caption{Visualize the feature attribution result of $W^{*} \cdot x$ to understand the model prediction.}
%\label{fig:imagenet_evaluation_result}
%\vspace{-2mm}
%\end{figure}

\subsection{Explainability of $W\cdot x$}
In previous section, we has discussed that decompose the prediction of neural network into 2 components.
In the datasets such as MNIST, $W \cdot x$ has similar prediction accuracy as the original model to explain the model prediction.
However, for dataset such as cifar10, cifar100 and ImageNet dataset, it does not show promising accuracy to explain the majority prediction of the neural network model.
During the model prediction, $b = \sum_{i=1}^{n-1}(W_{n}^{'}...W_{i+1}^{'} \cdot b_{i})+b_{n}$ plays a significant role to determine the prediction. 
It worth to notice that the convolution layer of convolution neural network model does not contain the bias term.
During the model decomposition, the bias term $b$ comes from the batch normalization layers.
For each value of $b_i$, its value is constant and determined once the model is trained.
In the equation, the variance that updated is $W'_i$ which is triggered by the activation of the different input. $W=W_{n}^{'} \cdot W_{n-1}^{'} ... W_{2}^{'} \cdot W_{1}^{'}$ contains the unique footprint of a model's reaction to a prediction.

An alternative approach to improve the sufficient of $W$ is to train a neural network without batch normalization layer.
Since the prediction of these networks can be represented as $F(x) = W \cdot x + b_n$ and $b_n$ is the bias term in the last layer of neural network.
We can use the $W$ to explain the network behavior directly.
%In Table~\ref{tab:modify_network}, we display the result of using common CNN architecture with modification to train these three popular dataset.
%For a simple dataset such as MNIST, we can end up with a model with high prediction accuracy of $W \cdot x$. 
A potential approach to improve the performance of such models include techniques such as weight normalization~\cite{10.5555/3157096.3157197},  initialization and other batch normalization free techniques.
Previous researches have demonstrate that network training without batch normalization can still achieve state-of-art performance~\cite{brock2021high}

%\section{Compare Feature Attribution}
\section{Applications}

In this section, we discuss the application of our proposed approach in supervise task and unsupervised task.

\begin{figure*}[t]
\centering
    \includegraphics[width=1\linewidth]{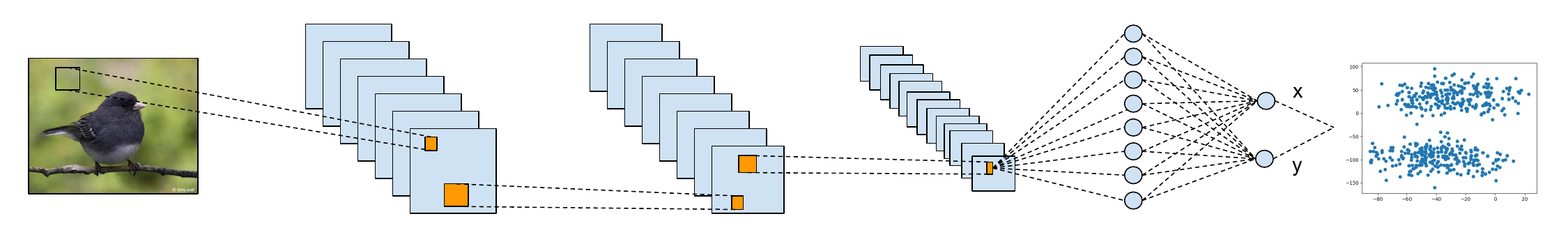}
   \vspace{-3mm}
    \caption{The neural network architecture that used to perform auto encoder training and parametric t-sne dimension reduction.}
\label{fig:background_architecture}
\vspace{-2mm}
\end{figure*}

\begin{figure*}[t]
\centering
    \includegraphics[width=\linewidth]{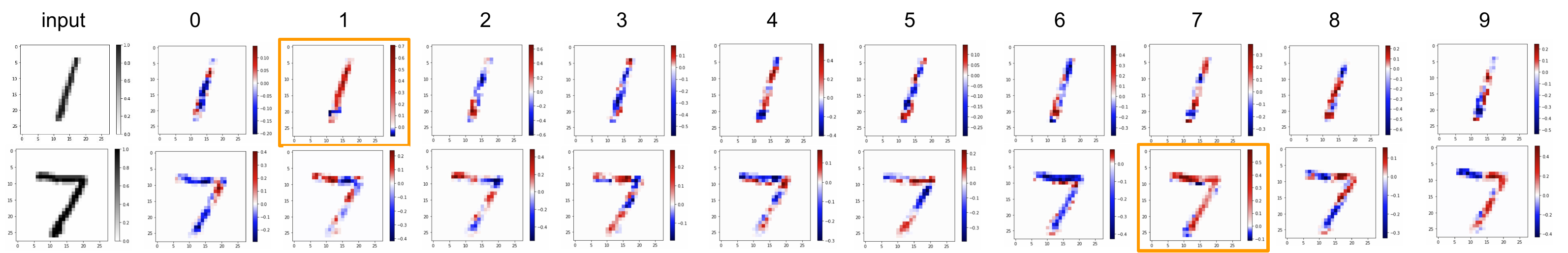}
    \vspace{-3mm}
    \caption{How neural network use input features to make a prediction in MNIST dataset with LeNet5.}
\label{fig:mnist_prediction}
\vspace{-2mm}
\end{figure*}

\begin{figure*}[t]
\centering
    \includegraphics[width=\linewidth]{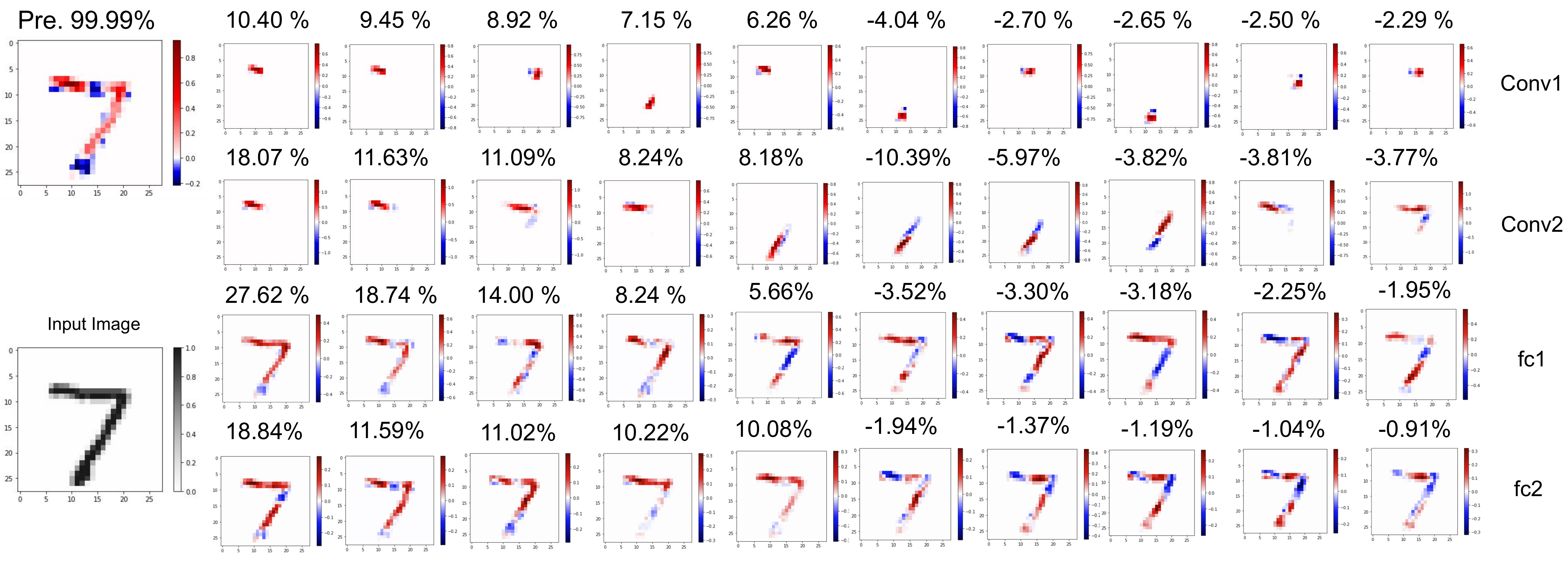}
    \vspace{-3mm}
    \caption{Displaying the features that captured by top 5 positive neurons and 5 negative neurons with their contribution to a model's final prediction in different layers of the LeNet5 model.}
\label{fig:mnist_activation_feature}
\vspace{-2mm}
\end{figure*}

\subsection{Supervised Learning - Image Prediction}
For MNIST dataset, we use our method to understand the decision of LeNet5, which the encoder layer does not include bias term.
The accuracy of the model is 0.9904.
For MNIST dataset, $W$ for the final prediction is a $(10,784)$ matrix and for each digit outcome, $W$ will give an explanation.
Each row of $W$ tells how a neural network use input feature to tell the prediction score for that digit.
Our feature attribution has natural semantic meaning.
The value, which assigned to the input pixel, multiplied with input pixel value is the value of current pixel contribute to the prediction.
A positive value has a positive contribution to the prediction and a negative value has a negative contribution.

Figure~\ref{fig:mnist_prediction} demonstrates how LeNet5 processes input digits image 1 and 7 for different label decision.
In the heat map visualization, the red color indicates the positive contribution and the blue color tells the negative contribution.
From the result, we can tell that both digits are recognized by the neural network model in the right shape.
It property of our approach is that compute the our feature attribution, it needs to compute the contribution of input to each neuron in the neural network first.
The final feature attribution is a summary of all previous' neuron's contribution.
This property brings the convenient to not only understand how neural network use input feature for decision but also provides how each network neuron use input feature to produce activation output.
In Figure~\ref{fig:mnist_activation_feature}, we display our feature attribution of digits 7 and how top 5 most activate neurons in different layers of LeNet5 use input features.

\subsection{Unsupervised Learning - Parametric Dimension Reduction}

Dimension reduction such as t-SNE is a popular approach to understand the structure of neural network model.
However, the classical t-SNE algorithm is non-parametric. 
Therefore, how the input is mapped into two dimensional space is unknown.
The uninterpretable process of t-SNE dimension reduction cause confusion and misleading during the analysis.
Comparing with t-SNE algorithm, parametric t-SNE perform similar operations by training a neural network model to perform the dimension reduction.
In this work, we mainly discuss the neural network architecture in Figure~\ref{fig:background_architecture} to perform the dimension reduction.
Many works~\cite{van2009learning, svantesson2023get,9973230} have developed to implement parametric t-SNE.

The loss function (equation (1)) that used to train parametric t-SNE try to minimize the probability distribution difference between the original high dimension data and the projected low dimension data.
\begin{equation}
   L(\theta) = \sum_{i\neq j} p_{ij} log\frac{p_{ij}}{q_{ij}}
\end{equation}

\begin{equation}
   p_{j|i} = \frac{exp(-||x_i-x_j||^2 /2\sigma_{i}^2)}{\sum_{k\neq i} exp(-||x_i-x_k||^2 /2\sigma_{i}^2)}
\end{equation}

\begin{equation}
   p_{ij} = \frac{p_{j|i}+p_{i|j}}{2N} 
\end{equation}

\begin{equation}
   q_{ij} = \frac{exp(-||f_{\theta}(x_i) -f_{\theta}(x_j)||^2 /2\sigma_{i}^2)}{\sum_{k\neq i} exp(-||f_{\theta}(x_i) - f_{\theta}(x_k)||^2 /2\sigma_{i}^2)}
\end{equation}

Because of the non-linear properties of the neural network model. 
Understanding the relationship between input and output of these networks trained with different loss function is difficult.
In the literature, limited work focus on understanding how input contribute to the final two dimension output.
Gradient is the common approach to perform feature attribution of the neural network model.
However, it is difficult to generate gradient for a single sample with the loss function that used to train neural network for parametric t-SNE. 
Furthermore, parametric t-SNE~\cite{van2009learning} may use neural network with an encoder with unsupervised approach to pretrain a network model, then use the neural network to perform the dimension reduction.
The overall process involve two neural network model which make the interpretability of the parametric t-SNE difficult.
Our approach is flexible enough to fill this gap by concatenate two network's matrix multiplication to understand the process of the parametric dimension reduction and the final projection can still be phased as $y= Wx+b$ for each sample's behavior.

\begin{figure}[t]
\centering
    \includegraphics[width=\linewidth]{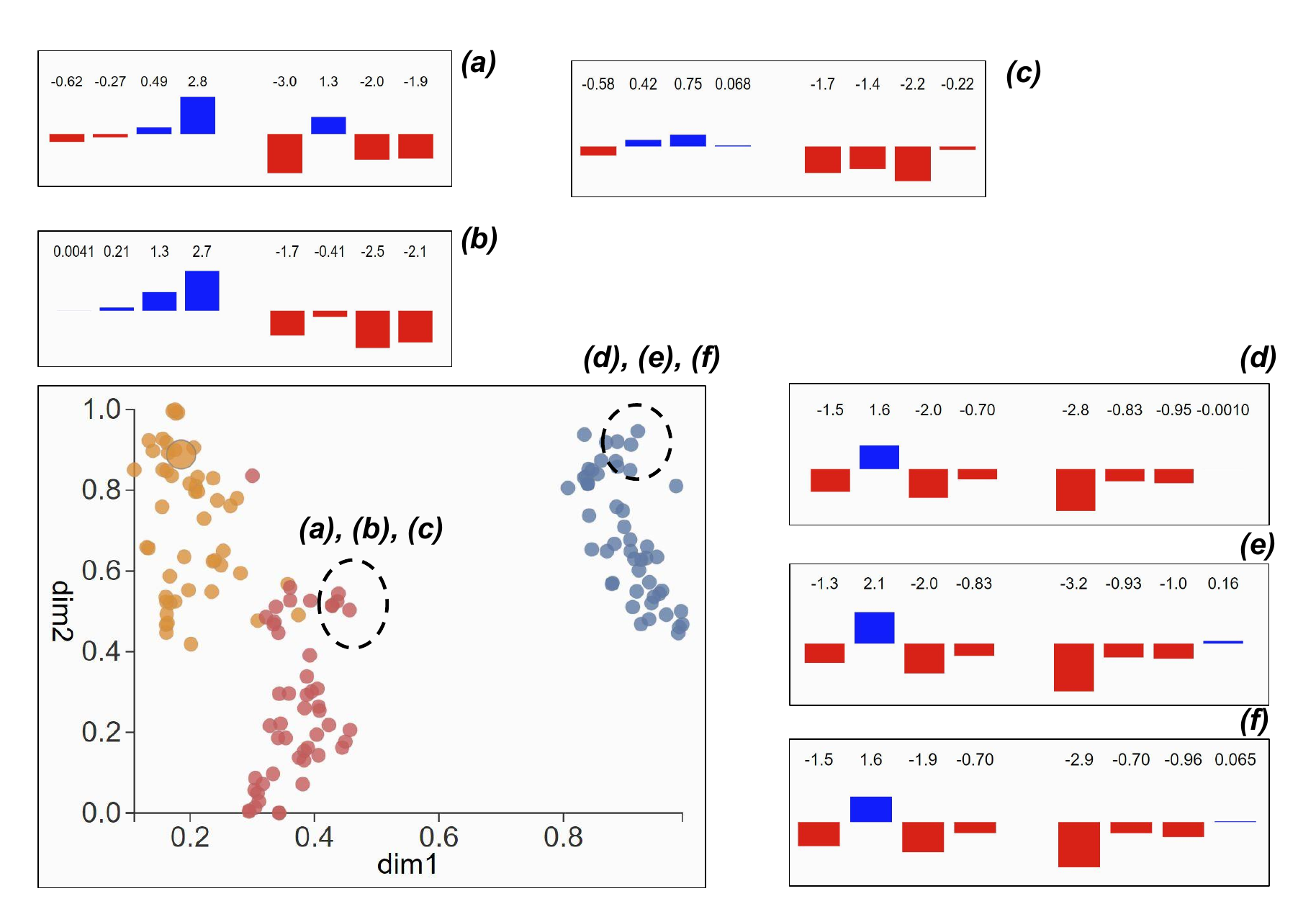}
   \vspace{-3mm}
    \caption{iris flower dataset is projected into two dimensional space with parametric t-SNE. With our feature attribution, it highlights the critical features that are used for dimension reduction.}
\label{fig:feature_attribution}
\vspace{-2mm}
\end{figure}

What features are used during dimension reduction process is important to interpret the dimension reduction result.
In Fig.~\ref{fig:feature_attribution}, we apply parametric t-SNE dimension reduction approach on iris dataset and use our proposed method to generate the feature attribution for dimension reduction result.
t-SNE dimension reduction often projects samples that are similar to each other to the nearby location.
From visualization, samples include (d), (e), (f) are nearby and the feature attribution result over these samples are similar.
However, samples (a), (b), (c) are nearby with very different feature attribution result.

%\subsection{Understand Error Propagation Through Neural Network Prediction}

%\subsection{Feature Attribution for Ensemble Model}

%\subsection{Samples' Feature Attribution stability over Margin}
\section{Conclusion}
In this work, we propose an approach to reformulating the forward propagation computation to understand a neural network prediction.
Our study find that a prediction of neural network can be rephrased as a series of matrix multiplication.
For each input instance, its output have a straight forward mapping which can be phrased as $y = W\cdot x+b$. 
At the end, we demonstrate the flexibility of our approach on how this approach can help use to understand the supervise classification task, and unsupervised dimension reduction task.

\section*{Acknowledgement}
This work was supported by  National Science Foundation Office of Advanced Cyber OCA-2138811.
This work was also performed under the auspices of the U.S. Department of Energy by Lawrence Livermore National Laboratory under Contract DE-AC52-07NA27344. The work is partially supported by Laboratory Directed Research and Development Program under tracking code 23-ERD-029. The work is reviewed and released under LLNL-TM-856619.

{\small
\bibliographystyle{ieee_fullname}
\bibliography{main}
}

\end{document}